\documentclass{article}
\usepackage{M-PSI}
\usepackage{amsmath,amssymb,amsfonts}
\usepackage{algorithmic}
\usepackage{graphicx}
\usepackage{textcomp}
\usepackage{booktabs}
\usepackage{subcaption}  
\usepackage{xcolor}
\usepackage{url}
\usepackage{multirow}
\usepackage{enumitem}
\usepackage[linesnumbered,ruled,vlined]{algorithm2e}
\def\BibTeX{{\rm B\kern-.05em{\sc i\kern-.025em b}\kern-.08em
    T\kern-.1667em\lower.7ex\hbox{E}\kern-.125emX}}

\usepackage[switch]{lineno}

\author{Xi Chen\textsuperscript{1,2}, Julien Cumin\textsuperscript{1}, Fano Ramparany\textsuperscript{1}, \href{https://research.vaufreydaz.org/}{Dominique~Vaufreydaz\textsuperscript{2,~\small{\ExternalLink }}}\vspace{0.1cm}\\
{$^1$ Orange Innovation}\\
{$^2$ Univ. Grenoble Alpes, CNRS, Grenoble INP, LIG, 38000 Grenoble, France}\\ 
}

\title{\textbf{MuRAL}: A Multi-Resident Ambient Sensor Dataset Annotated with Natural Language for Activities of Daily Living}

\usepackage[pdfencoding=auto, pdfusetitle]{hyperref}
\hypersetup{
  hidelinks,
  urlcolor=red,
  pdftitle={MuRAL: A Multi-Resident Ambient Sensor Dataset Annotated with Natural Language for Activities of Daily Living},
  pdfauthor={Xi Chen, Julien Cumin, Fano Ramparany, Dominique Vaufreydaz},
  pdfkeywords={Human Activity Recognition, Large Language Models, Ambient Intelligence, Smart Homes.},
}

\author{Xi Chen\textsuperscript{1,2}, Julien Cumin\textsuperscript{1}, Fano Ramparany\textsuperscript{1}, \href{https://research.vaufreydaz.org/}{Dominique~Vaufreydaz\textsuperscript{2,~\small{\ExternalLink }}}\vspace{0.1cm}\\
{$^1$ Orange Research}\\
{$^2$ Univ. Grenoble Alpes, CNRS, Grenoble INP, LIG, 38000 Grenoble, France}\\ 
}

\removefirstpagelogo{}
\begin{document}

\begin{abstract}
Recent progress in Large Language Models (LLMs) has enabled advanced reasoning and zero-shot recognition for human activity understanding with ambient sensor data. However, widely used multi-resident datasets such as CASAS, ARAS, and MARBLE lack natural language context and fine-grained annotation, limiting the full exploitation of LLM capabilities in realistic smart environments. To address this gap, we present \textbf{MuRAL} (\underline{Mu}lti-\underline{R}esident \underline{A}mbient sensor dataset with natural \underline{L}anguage), comprising over 21 hours of multi-user sensor data from 21 sessions in a smart home. MuRAL uniquely features detailed natural language descriptions, explicit resident identities, and rich activity labels, all situated in complex, dynamic, multi-resident scenarios. We benchmark state-of-the-art LLMs on MuRAL for three core tasks: subject assignment, action description, and activity classification. Results show that current LLMs still face major challenges on MuRAL, especially in maintaining accurate resident assignment over long sequences, generating precise action descriptions, and effectively integrating context for activity prediction. The dataset is publicly available at: \href{https://mural.imag.fr}{https://mural.imag.fr/}.

\end{abstract}

\section{Introduction}

The recognition of Activities of Daily Living (ADL) in smart home environments plays a crucial role in various applications such as healthcare \cite{acampora2013survey, riboni2016smartfaber}, security \cite{tripathi2018suspicious, dahmen2017activity}, and home automation \cite{thomas2016activity}. Compared to camera-based approaches, ambient sensors such as magnetic contact sensors, movement sensors, and smart plugs offer advantages in terms of privacy, energy efficiency, and non-intrusiveness. Unlike wearable sensors, they do not require users to wear devices continuously, making them more comfortable and user-friendly. These properties have driven increasing interest in ambient-sensor-based ADL recognition in recent years.

A unique challenge of ambient sensor data lies in its low-level, discrete nature. For example, a door sensor typically outputs only “open” or “closed” states. The semantic meaning of such readings is highly context-dependent: opening a cabinet, a room door, or a front door may imply very different intentions. This context dependency makes accurate interpretation difficult for conventional machine learning models, which often lack access to semantic context. In contrast, large language models (LLMs) are well-suited to this task when provided with rich natural language descriptions of sensor context~\cite{thukral2025layout, chen2024towards}. Recent studies have thus explored LLM-based approaches for interpreting ambient sensor data through zero-shot reasoning.

Leveraging their powerful in-context learning abilities, LLMs have recently achieved significant breakthroughs in ambient-sensor-based ADL recognition. Most studies~\cite{gao2024unsupervised, sun2024ai, cleland2024leveraging, shoumi2024leveraging, ugwu2025potential, thukral2025layout, civitarese2024large} formulate sensor inputs with environmental context as natural language prompts, enabling LLMs to recognize activities under zero-shot or few-shot settings in single-resident scenarios. Chen et al.~\cite{chen2024towards} extended this paradigm to multi-resident environments~\cite{arrotta2024multi}, using chain-of-thought (CoT) \cite{wei2022chain} reasoning to disambiguate mixed sensor streams and provide fine-grained descriptions for each resident. Other works exploit LLMs for activity generation~\cite{takeda2023sensor, yonekura2024generating}, explainability~\cite{fiori2025leveraging}, or even raise privacy concerns~\cite{juttner2025chatanalysis} in smart home contexts.

However, most existing studies in this field still rely on traditional benchmark datasets that were developed prior to the emergence of LLM, such as CASAS \cite{cook2012casas}, ARAS \cite{alemdar2013aras}, and MARBLE \cite{arrotta2021marble}. While these datasets have played a crucial role in advancing ambient-based ADL recognition over the past decade, they exhibit several limitations when applied in the context of LLM-based approaches. Specifically, traditional datasets typically lack rich contextual metadata, are collected in relatively simple settings with scripted routines, and focus primarily on high-level activity classification rather than fine-grained textual description tasks. These limitations collectively hinder the full exploration of LLM potential for ambient-based ADL recognition.

To bridge this gap, we introduce \textbf{MuRAL}, a \textbf{Mu}lti-\textbf{R}esident \textbf{A}mbient sensor dataset with natural \textbf{L}anguage, specifically designed to unlock the full reasoning capabilities of LLMs for smart-home activity understanding. Unlike existing datasets, MuRAL provides fine-grained natural language descriptions for sensor events, captures complex and dynamic multi-resident scenarios with 2-4 residents, and includes rich contextual annotations at both action and session levels. These characteristics make MuRAL particularly valuable for modeling ambiguous, overlapping, and socially situated activities—challenges that are rarely addressed in existing datasets~\cite{arrotta2024multi} but essential for next-generation intelligent activity recognition systems.

In the remainder of this paper, Section~\ref{sec: collection} describes our data collection environment, annotation methodology, and provides an overview of MuRAL. Section~\ref{sec: experiments} presents benchmark evaluations. The dataset is publicly available at: \href{https://mural.imag.fr}{https://mural.imag.fr/}.

\section{MuRAL Dataset} \label{sec: collection}

\begin{figure}[t]
    \centering
    \begin{subfigure}[b]{0.23\textwidth}
        \centering
        \includegraphics[width=\linewidth]{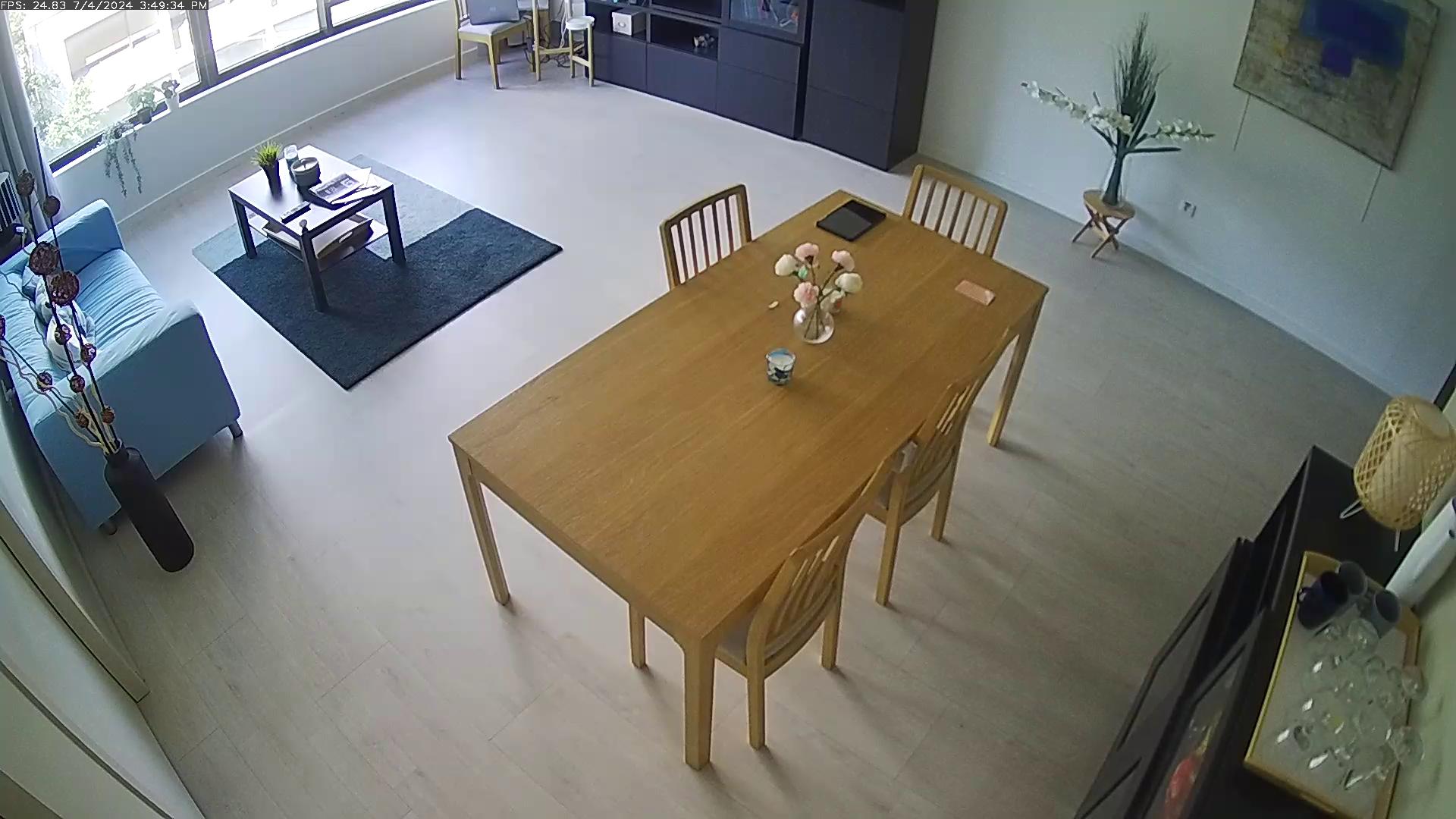}
        \caption{Living room and dining room}
    \end{subfigure}
    \hspace{0.1em}
    \begin{subfigure}[b]{0.23\textwidth}
        \centering
        \includegraphics[width=\linewidth]{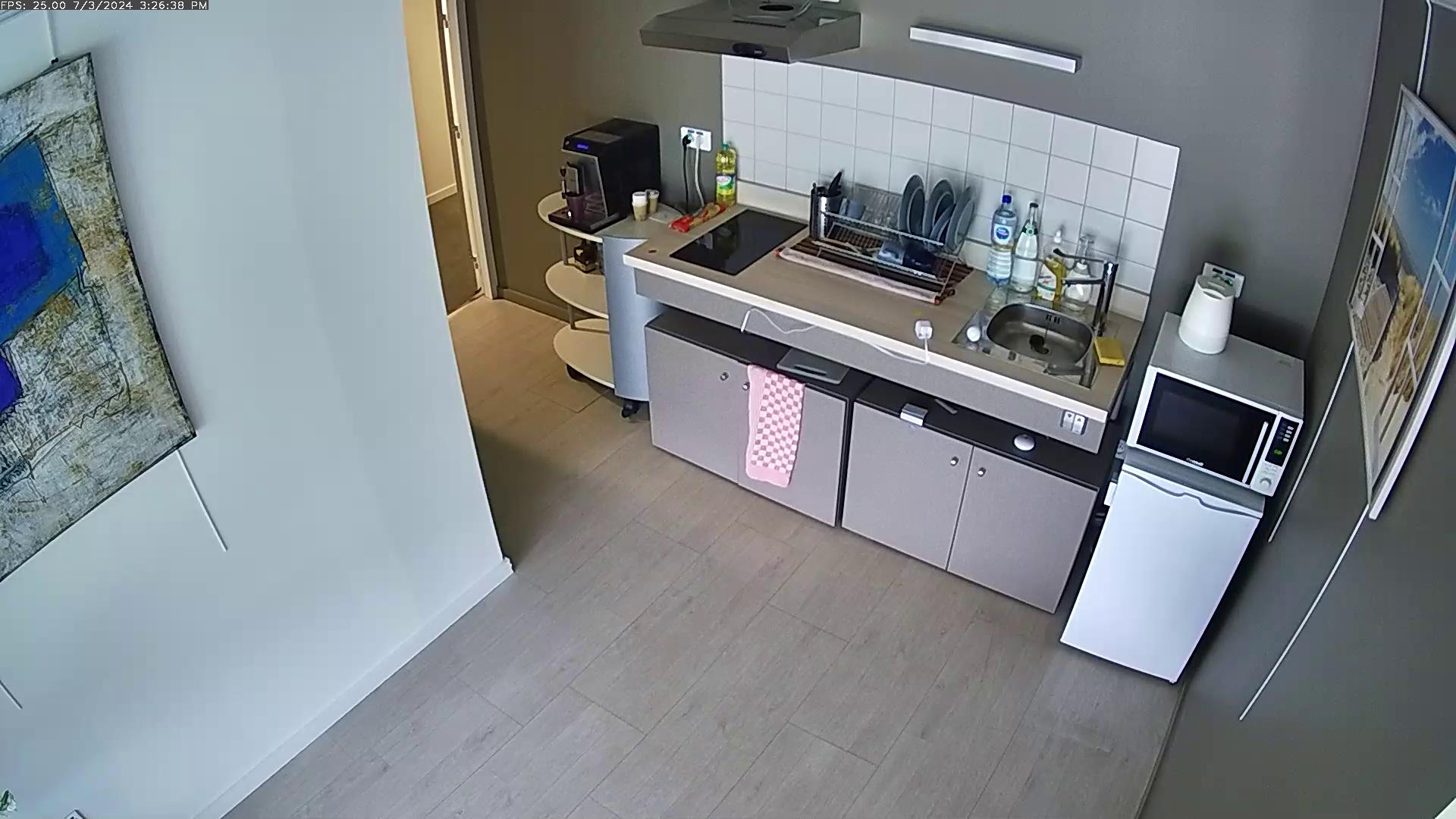}
        \caption{Kitchen}
    \end{subfigure}
    
    \vspace{0.5em}
    \begin{subfigure}[b]{0.23\textwidth}
        \centering
        \includegraphics[width=\linewidth]{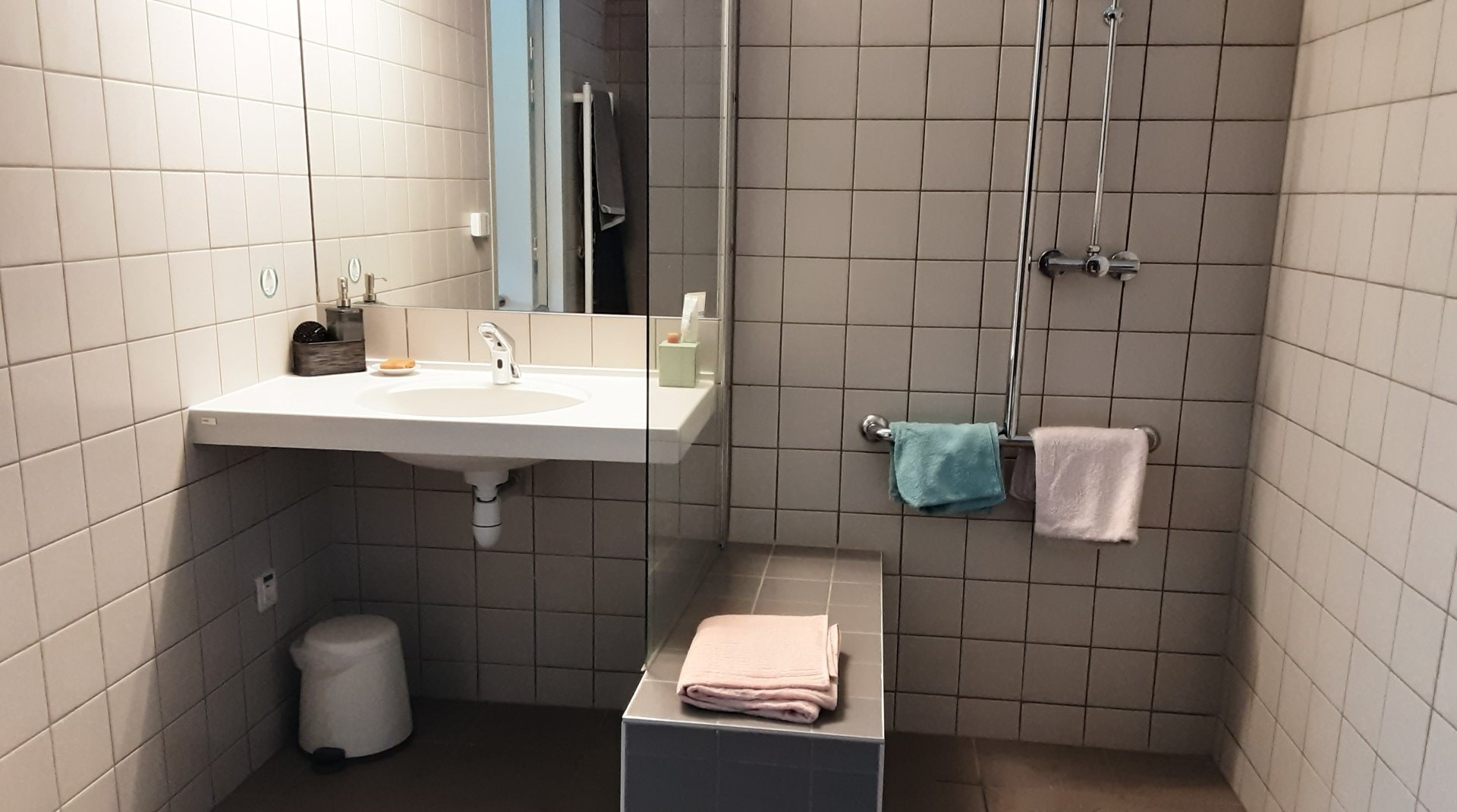}
        \caption{Bathroom}
    \end{subfigure}
    \hspace{0.1em}
    \begin{subfigure}[b]{0.23\textwidth}
        \centering
        \includegraphics[width=\linewidth]{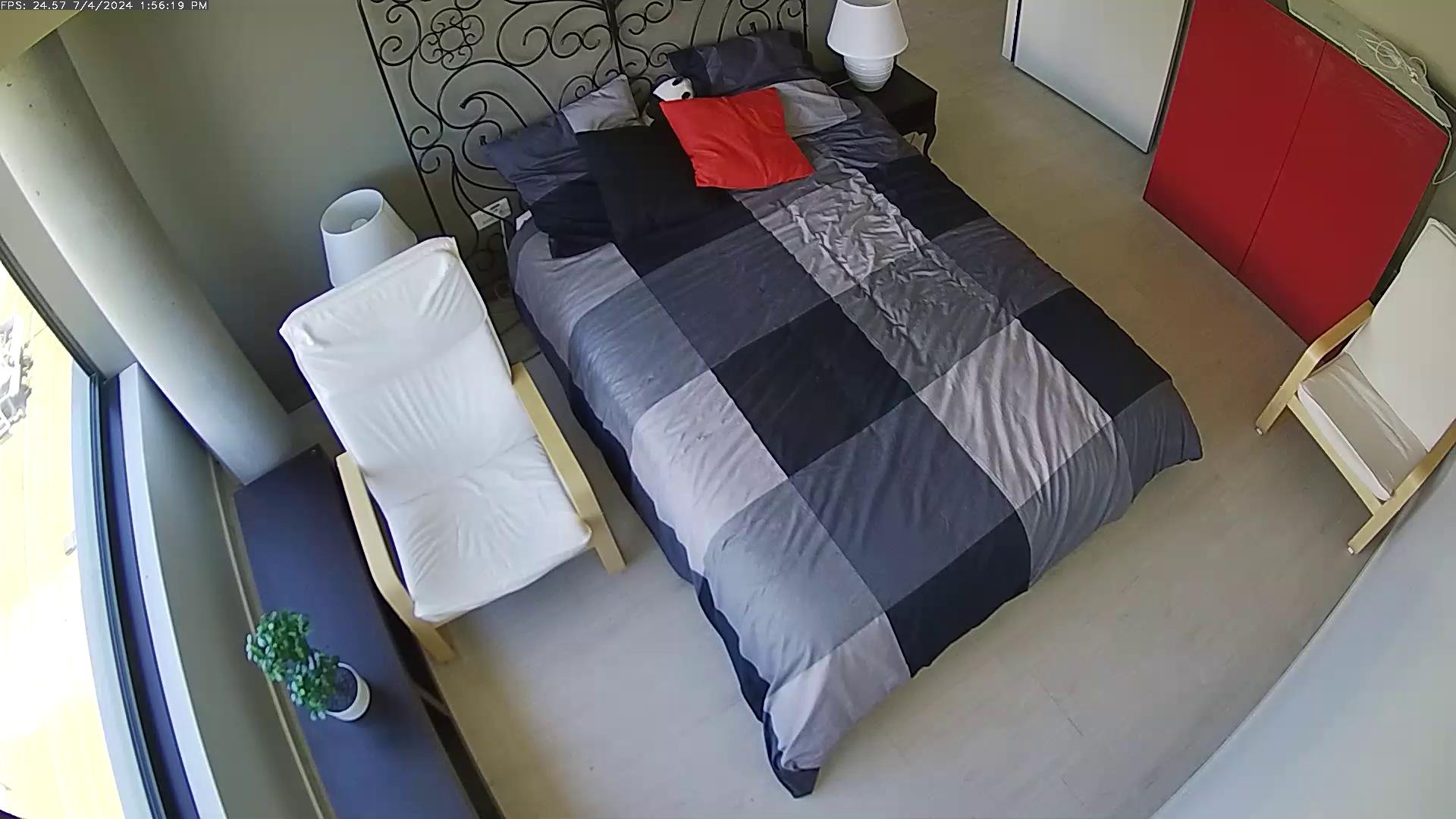}
        \caption{Bedroom 1}
    \end{subfigure}

    \caption{DOMUS intelligent apartment experimental platform. Photo credit: withheld for blind review.}
\label{fig:domus}
\end{figure}

\subsection{Environmental Setup}

The MuRAL dataset was collected in the DOMUS\footnote{\url{https://www.liglab.fr/fr/recherche/plateformes/domus}} Intelligent Apartment, a 60 m$^2$ living lab designed to simulate real-world residential environments. As shown in Figure \ref{fig:domus}, the apartment includes a kitchen (with a coffee maker, induction stove, cabinets, sink, countertop, microwave, and refrigerator), a dining area (dining table, chairs, and snack cabinet), a living room (couch, coffee table, TV, gaming console, and bookshelf), a bathroom (shower, toilet, and sink), as well as one double and one single bedroom.

\begin{figure}[t]
    \centering
    \includegraphics[width=0.97\linewidth]{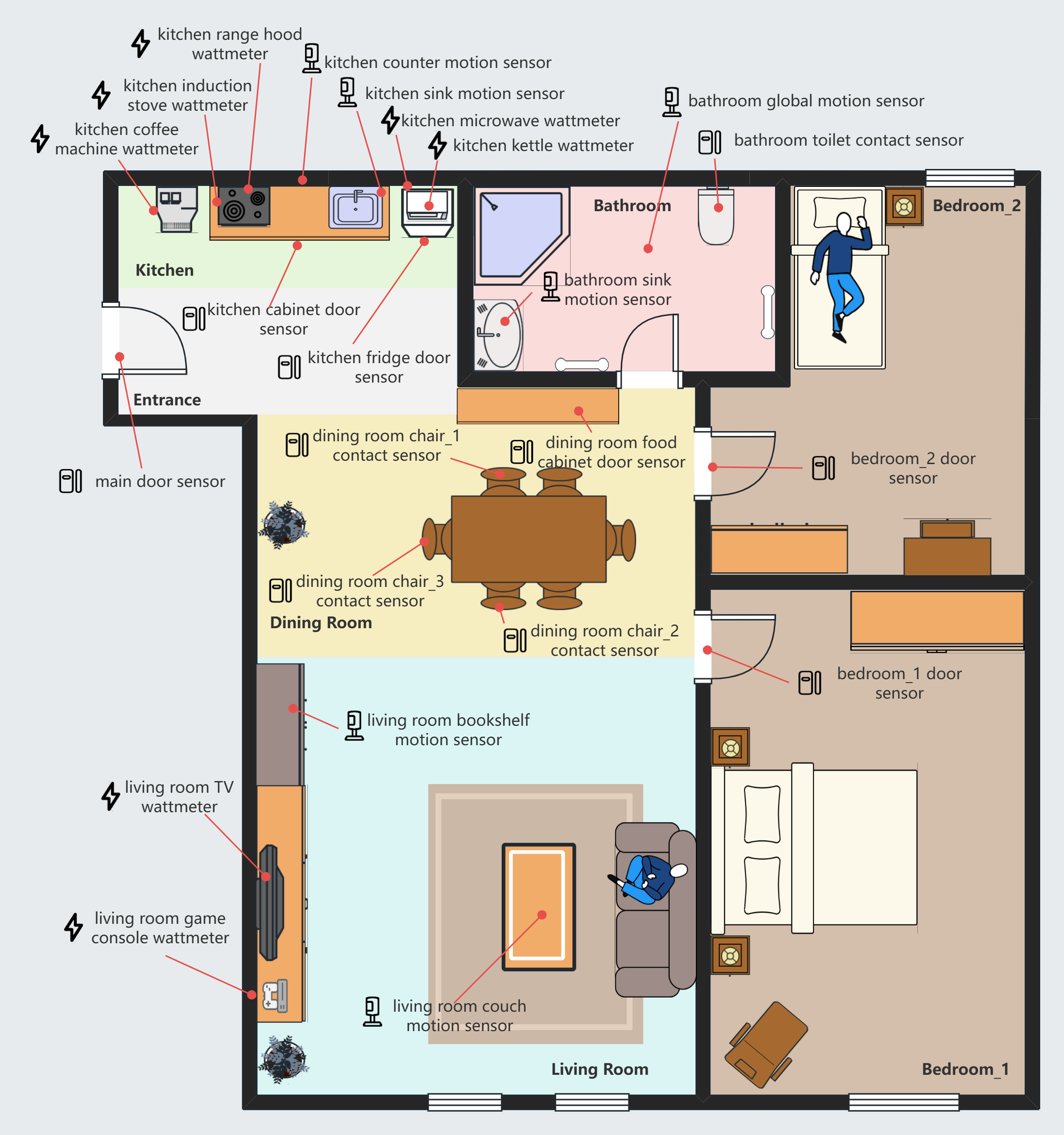}
\caption{Floor plan of the DOMUS intelligent apartment with the names and locations of all ambient sensors.}
\label{fig:layout}
\end{figure}
The apartment is equipped with 23 ambient sensors (Figure~\ref{fig:layout}), including 6 infrared motion sensors (Fibaro FGMS001\footnote{\url{https://www.openhab.org/addons/bindings/zwave/thing.html?manufacturer=fibaro&file=fgms001_3_2.html}}, AEON ZW100\footnote{\url{https://www.openhab.org/addons/bindings/zwave/thing.html?manufacturer=aeon&file=zw100_0_0.html}}), 10 magnetic contact sensors (Philio PSM02\footnote{\url{https://www.openhab.org/addons/bindings/zwave/thing.html?manufacturer=philio&file=psm02_0_0.html}}), and 7 smart plugs (Fibaro FGWP102\footnote{\url{https://www.openhab.org/addons/bindings/zwave/thing.html?manufacturer=fibaro&file=fgwp102_3_2.html}}) to monitor the power consumption of the appliance. 
These sensors were deployed such that their sensing ranges do not overlap, and each sensor is assigned to a dedicated and well-defined activity zone (such as a sink or a bookshelf). This configuration helps ensure that sensor events correspond clearly to specific user actions or contexts, thus facilitating the interpretability of data for LLMs and improving downstream activity recognition. Furthermore, the sensitivity of all sensors was set to 1 Hz to ensure timely reflection of residents' activities and to prevent consecutive triggers of the same sensor by different residents from being merged into a single event. To enable the eventization of smart plug data, we preset thresholds for the power and rate of change for each appliance.

\begin{table*}[t]
  \centering
  \caption{Activity ID and labels of the MuRAL Dataset.}
  \label{tab:activity_id}
  \resizebox{\textwidth}{!}{  
\begin{tabular}{lllll}
\hline
0. others                              & 1. using toilet             & 2. preparing drinks              & 3. preparing breakfast         & 4. having breakfast           \\
5. grabing food or drinks              & 6. bedroom personal activity & 7. grabing tableware             & 8. clearing table              & 9. watching TV                \\
10. drinking water, coffee or dinks   & 11. cleaning appartment     & 12. taking shower                & 13. washing dishes             & 14. getting out of the apartment \\
15. coming into the apartment         & 16. reading books            & 17. playing video games          & 18. resting on the couch or the chairs & 19. preparing dinner \\
20. having dinner                     & 21. chatting                 & 22. playing board games          & 23. treating fruits            & 24. having fruits or snacks  \\
25. personal washing                  & 26. using pc                &                                  &                                &                              \\
\hline
\end{tabular}
  }
\end{table*}

\begin{figure*}[t]
    \centering
    \includegraphics[width=\linewidth]{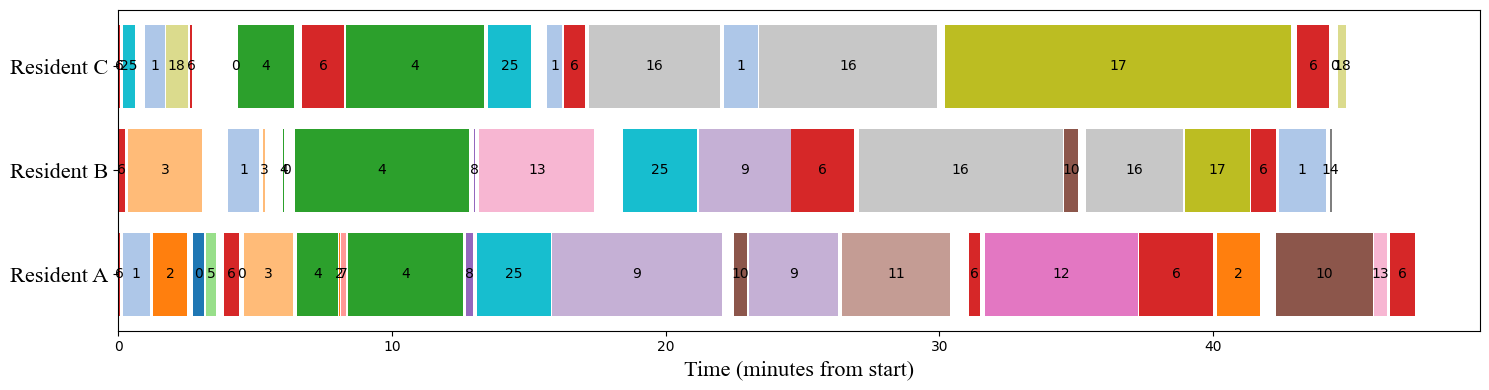}
\caption{Activity timeline for multiple residents in Session 01 of MuRAL.}
\label{fig:timeline}
\end{figure*}

However, considering privacy concerns in real-world scenarios, no sensor is installed inside the bedrooms, leaving bedroom activities outside of the main scope of our study. All activities in bedrooms are annotated as ``bedroom personal activity''. Instead, magnetic sensors are placed on the bedroom doors to indicate the entry and exit of residents. In addition, due to the lack of pressure sensors, alternative sensing methods were employed to infer interactions with certain furniture. For example, magnetic sensors were installed between the dining table and chairs to detect whether a chair was in use, requiring participants to return the chair to its original position after use to reset the sensor. Similarly, a magnetic sensor is installed on the toilet lid to determine whether it was open or closed, instructing participants to lower the lid after each use to ensure consistent event recording. For couch activity detection, an motion sensor is placed on the coffee table and limited its detection range to the couch area, allowing us to infer whether someone was present. It is important to note that these compromising adaptations may introduce slight deviations from natural behavior.

In addition to the ambient sensors used to collect the resident activity data, we also used the six ceiling-mounted cameras and their associated microphones pre-installed in the common areas (kitchen, living room, and dining area) to record video and audio. These video/audio recordings were used solely for annotation purposes with the consent of participants and were permanently deleted as soon as the data annotation process was complete.

In terms of software, DOMUS uses the \textit{OpenHAB platform\footnote{\url{https://www.openhab.org/}}} to connect, control, and integrate all the sensors, and finally collect all sensor data. This is an open-source smart home platform designed to integrate and automate IoT devices from different manufacturers and protocols. When a sensor detects a state change or generates a new reading, it transmits a message containing the timestamp of the event or reading, the sensor ID, and the corresponding data to the OpenHAB platform. The platform collects messages from all sensors and organizes them into a time-ordered data log based on the associated timestamps.

\subsection{Data Collection}

MuRAL includes data collected from 18 volunteer participants. To ensure privacy and fairness, no information regarding participants' age, occupation, nationality, or gender was recorded, selected, or analyzed. No minors were involved. Volunteers were recruited via internal university mailing lists, making it reasonable to assume that most participants were students, researchers, or educators, while other demographic attributes remained diverse. Similarly, no explicit identifiers were recorded during data collection, and re-identifying individuals from the dataset is considered to be impossible. Participants were scheduled across 21 sessions based on their self-selected time slots, with each session consisting of 2 to 4 people. Across all sessions, only session 06 included 2 participants, session 05 had 4 participants, while the remaining 19 sessions had 3 participants.

To enable LLMs to effectively reason about complex real-world daily activities, it is essential that the data distribution closely mirrors natural human behavior. This requires that scene configurations, role assignments, interactions, activity sequences, and durations during data collection are representative of authentic household routines. In our data collection protocol, participants were assigned specific social roles (e.g., family members, spouses, roommates) and predefined temporal contexts (e.g., weekday morning, weekday evening, weekend afternoon). Rather than prescribing fixed activity sequences, we provided only high-level contextual information, such as role descriptions, time-of-day framing, and a list of possible activities. Participants were then encouraged to act freely based on their own habits and preferences. This minimally guided approach promoted spontaneous and diverse interactions while preserving realism in behavior patterns.

To support the execution of naturalistic activities, the environment was carefully prepared. Prior to data collection, participants were introduced to the apartment layout, briefed on how to operate household appliances, and given 10 to 20 minutes to explore and familiarize themselves with the space. The apartment was stocked with typical household items: the refrigerator contained ingredients such as pasta, eggs, salad, microwave meals, fruits, and ice cream; the snack cabinet included chips, cookies, and chocolate; the bookshelf offered a variety of books and board games; and the gaming console provided access to several casual video games. These resources enabled participants to engage in realistic sequences of activities—such as cooking, dining, reading, or relaxing—thereby producing sensor data grounded in genuine user intentions.

To better reflect natural social dynamics, participants were permitted to enter or leave the apartment over the course of the session as the scenario evolved. Consequently, the number of active residents varied over time. Each session lasted approximately one hour, allowing for the capture of multiple overlapping activities and temporal transitions. This setting yielded rich, temporally structured data that supports downstream reasoning and activity modeling tasks. For privacy and hygiene considerations, personal care activities (e.g., showering, brushing teeth, using the toilet) were simulated rather than executed. In contrast, other activities, such as cooking and dining, were performed naturally, contributing to the authenticity and ecological validity of the dataset.

\subsection{Data Annotation}

Unlike former datasets~\cite{arrotta2021marble,cook2012casas,alemdar2013aras}, MuRAL offers fine-grained natural language descriptions of each participant's actions within long, complex multi-resident scenarios, significantly enhancing the annotation difficulty. To ensure accuracy and consistency, all annotations were completed by a single annotator based on the recorded video footage from each session. By synchronizing the timestamps of each video frame with those of the collected sensor data, annotators were able to determine labeling information by observing the participants’ identities, actions, and activities in the video.

To annotate the residents associated with each event, the annotator assigned alphabetical identifiers to participants, following the order of their appearance in the session and adhering to alphabetical sequencing. For example, the resident who triggered the first event in the session was assigned an identifier ``\textbf{A}'', the next distinct resident who triggered a non-A event was assigned ``\textbf{B}'', and so on. 

After annotating the identities of the residents, the annotator then wrote a natural language description of the resident’s action and the context of the event. For example, a ``\texttt{kitchen cabinet door CLOSED}'' event might be annotated as:
\textit{``A takes out dishes from the kitchen cabinet, moves them to the dining room, and places the dishes on the table.''}. It is worth noting that not every event was annotated with a description. If the resident’s current activity showed no significant change from the previously annotated description, annotators skipped the description labeling for that event. According to our statistical analysis shown in Table~\ref{tab:transposed_statistic}, about half of the events are described.

In addition to fine-grained action descriptions, we also provide higher-level activity labels, similar to previous datasets~\cite{arrotta2021marble,cook2012casas,alemdar2013aras}, enabling the dataset to support activity recognition classification tasks. After completing the identity and action annotations, the annotator grouped events by resident identifier and labeled each resident’s sequence with one of 27 predefined activity categories, based on both their action descriptions and the recorded video. These activities are listed and presented in Table \ref{tab:activity_id}. Figure \ref{fig:timeline} provides an example of an activity timeline depicted based on these annotations.

\subsection{Dataset Overview}
The collected dataset is organized by session, with each session stored in a separate folder. Inside each folder, there is a CSV file containing sensor event records and a JSON file that provides contextual information about the session. The CSV file includes 7 columns: the unique identifier of the event (\texttt{uid}), the time of the day for the event (\texttt{time}), the name of the triggered sensor (\texttt{sensor}), the state change of sensor (\texttt{action}), the identifier of the resident who triggered the event (\texttt{subject}), a textual description of the resident’s action (\texttt{description}), and the corresponding high-level activity label (\texttt{activity}). The associated context file (\texttt{context.json}) captures metadata about the session, including the start time, the day type (e.g., weekday or weekend), the number of residents involved, their respective roles in the scenario, and the overall simulated household context.

In addition to the session folders, the dataset includes two JSON files at the top level: \texttt{activity.json}, which maps activity IDs to their semantic descriptions, and \texttt{sensor.json}, which provides detailed information about each sensor, including its type, location, and functional description. Furthermore, a file named \texttt{flootplan.png} visually illustrates the positions of all sensors within the DOMUS environment. This structure enables comprehensive and multi-level analysis of user behavior in multi-resident smart home environments, supporting tasks ranging from low-level event modeling to high-level activity reasoning.

\begin{table}[t]
\centering
\caption{Session-wise Statistics of the MuRAL Dataset.}
\label{tab:transposed_statistic}
\resizebox{0.45\textwidth}{!}{
\begin{tabular}{c|cccc}
\hline
Statistic & mean & std & min & max \\ \hline
\#Users               & 3.00 & 0.31 & 2    & 4    \\
\#Events              & 406.81 & 72.86 & 266  & 558  \\
\#Sensors             & 18.76 & 2.18  & 15   & 22   \\
\#Description         & 200.33 & 53.57 & 128  & 284  \\
Description Length    & 11.67 & 0.67  & 9.84 & 13.22\\
\#Activity            & 41.71 & 10.96 & 20   & 63   \\
\#Types of Activity   & 15.71 & 2.21  & 10   & 19   \\
Duration per Activity (min) & 3.75 & 1.35  & 1.98 & 7.01 \\ \hline
\end{tabular}
}
\end{table}

A statistical analysis of the MuRAL dataset is reported in Table~\ref{tab:transposed_statistic}. The reported statistics highlight its suitability for multi-resident activity recognition with natural language supervision. Across 21 sessions, each session contains on average 200 natural language annotations, with consistently fine-grained descriptions averaging 11.67 words in length. These annotations provide semantically rich insights into activities, sensor interactions, and resident behaviors. On average, each session includes 41.7 activity instances spanning 15.7 distinct activity classes, reflecting diverse activity coverage. The dataset also captures realistic multi-resident dynamics, with most sessions involving three residents and an average of 407 sensor events per session. Activity durations vary substantially (1.98 to 7.01 minutes), highlighting temporal diversity. With 18.76 sensors used per session out of a total of 23, the dataset spans a broad range of spatial and functional contexts, making it a strong benchmark for evaluating complex reasoning in smart home environments.

\section{Experimental Evaluation} \label{sec: experiments}

In this section, MuRAL is benchmarked using the state-of-the-art LLMs as a baseline for future potential work. We evaluate the results of this benchmark from three perspectives: resident assignment, action description, and activity classification.

\subsection{Approaches}

To the best of our knowledge, LAHAR~\cite{chen2024towards} represents one of the state-of-the-art approaches for multi-resident ADL recognition and is therefore selected as our primary baseline. However, LAHAR employs a pre-processing step that merges adjacent sensor events with identical sensor names, thereby increasing the input information density. This approach introduces two major limitations in the complex and dynamic environments targeted by MuRAL. First, in these settings, adjacent events from the same sensor may be triggered by different residents; merging such events risks conflating distinct actions and losing critical resident-specific information. Second, this event-merging strategy prevents the generation of action descriptions for individual sensor events, making it impossible to directly evaluate LLM-generated descriptions against MuRAL’s fine-grained, event-level annotations.

To address these limitations, we avoid pre-merging and instead treat each sensor event as an independent unit. Except for this modification, our pipeline closely follows that of LAHAR. Specifically, our framework operates in two stages. In the first stage, the event stream is segmented into blocks of 10 events, each formatted as a JSON array containing the timestamp, sensor metadata, and a sequence index. These blocks are processed by an LLM prompted to (i) assign each event to a resident, (ii) describe the likely action, and (iii) update the state for each resident. To ensure consistency across blocks, the final resident states from one block are provided as context to the subsequent block. In the second stage, resident assignments from the first stage are used to partition events by resident. Each resident’s event sequence is again divided into blocks of 10, with each block—containing the sequence index, timestamp, and action description—fed into a second LLM. This model is prompted to infer high-level activities from the action descriptions. The resulting outputs are mapped back to the original events using sequence indices, thus ensuring a one-to-one correspondence and enabling both qualitative and quantitative evaluation.

\subsection{Metrics}
\paragraph{Resident Assignment Accuracy}

For long sequences with hundreds of events, enforcing a single global one-to-one mapping between ground truth and predicted resident labels can be unreliable, as early mismatches may propagate and distort the evaluation. To address this, each session is divided into shorter windows of size $K$ and evaluated independently. Within each window, resident labels in both ground truth and prediction sequences are reordered by their order of appearance and mapped to alphabetical order for consistency. For example, if the predicted sequence is ["C, B", "A", "C"], it is remapped to ["A, B", "C", "A"]. For each event in the window, accuracy is determined by comparing the sets of resident labels: a perfect match scores 1; if the predicted set has more residents than ground truth, or there are no matches, the score is 0; for partial matches, the score is computed as the number of matching residents divided by the total number of ground truth residents. The final resident assignment accuracy is defined as the average score over all events. In our experiments, we vary the window size $K$ from 2 to 256 to analyze its impact on evaluation robustness.

\paragraph{Cosine Similarity Score}

To evaluate the difference between the action descriptions \(\hat{A}\) generated by the LLM and the human-annotated descriptions \(A\), we use an embedding model to compute the embedding vectors \(\text{emb}(A)\) and \(\text{emb}(\hat{A})\) for each pair \((A, \hat{A})\).

We then compute the cosine similarity between the two embedding vectors as a measure of their semantic similarity:
\begin{align*}
\text{sim}(A, \hat{A}) 
&= \frac{ \text{emb}(A) \cdot \text{emb}(\hat{A}) }{ \| \text{emb}(A) \| \cdot \| \text{emb}(\hat{A}) \| }.
\end{align*}

Let a session contain \(N\) action-description pairs \((A_i, \hat{A}_i)\), where \(i = 1, 2, \ldots, N\). The overall action description score for the session is computed as the average cosine similarity.

\paragraph{Accuracy}
For the high-level activity classification task, we adopt the standard classification accuracy.

\subsection{Results}


\begin{table}[t]
\caption{Comparisons of model performances across 3 tasks.}
\label{tab: results}
 \resizebox{0.47\textwidth}{!}{
\begin{tabular}{l|l|c|c|c}
\hline
Method                 & Model   & Assignment             & Description            & Classification           \\ \hline
\multirow{2}{*}{LAHAR~\cite{chen2024towards}} & 4o      & 0.691          & 0.463          & 0.413          \\
                       & 4o-mini & 0.692          & 0.465          & 0.231          \\ \hline
\multirow{2}{*}{Ours}  & 4o      & \textbf{0.701} & 0.563          & \textbf{0.417} \\
                       & 4o-mini & 0.608          & \textbf{0.564} & 0.186          \\ \hline
\end{tabular}
}
\end{table}

Experiments are conducted on the MuRAL dataset, comparing LAHAR~\cite{chen2024towards} with our adapted approach as described above. The results were evaluated using the previously introduced metrics. In the experiments, we primarily used two models: GPT-4o and GPT-4o-mini~\cite{hurst2024gpt}, which are referred to as 4o and 4o-mini, respectively, in the following discussion. The results averaged across all sessions are summarized in Table \ref{tab: results}. For Resident Assignment, the accuracy is reported using a window size $K$ of 8.

\begin{figure}[t]
    \centering
    \includegraphics[width=\linewidth]{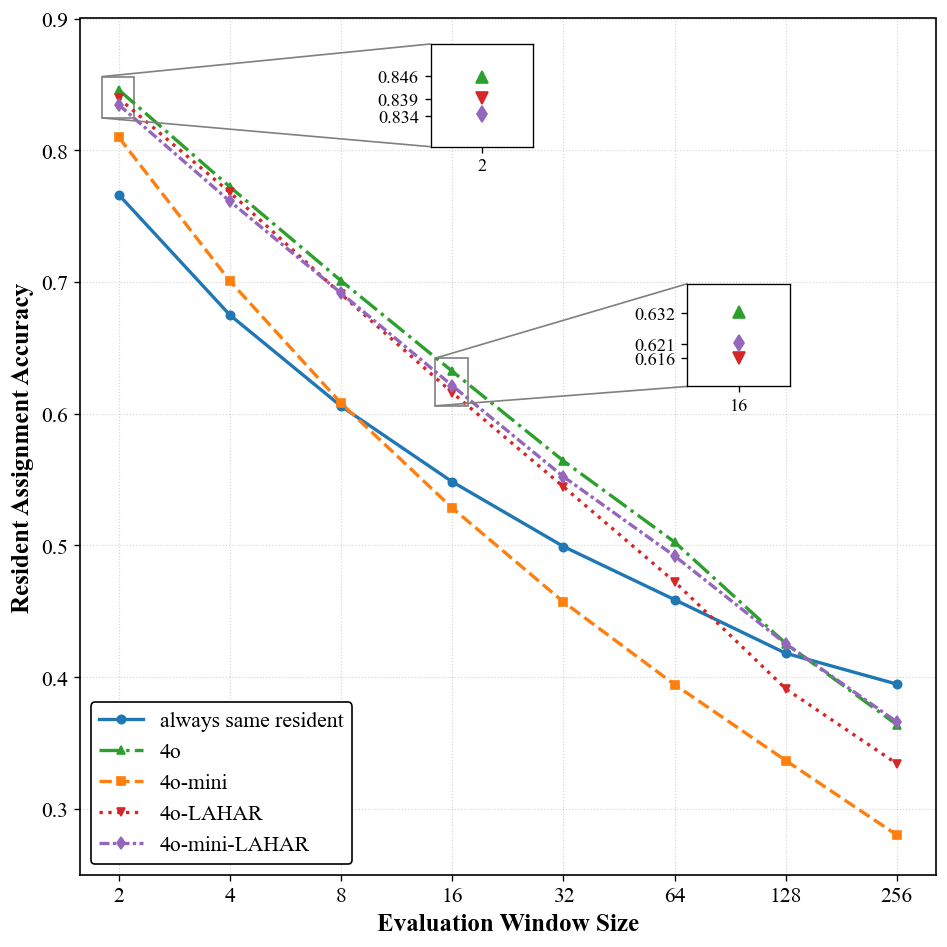}

    \caption{Comparison of assignment accuracy across different models as a function of evaluation window size (x-axis is log-scaled).}
    \label{fig: assign_vs_wins}
\end{figure}

\begin{figure}[t]
    \centering
    \includegraphics[width=\linewidth]{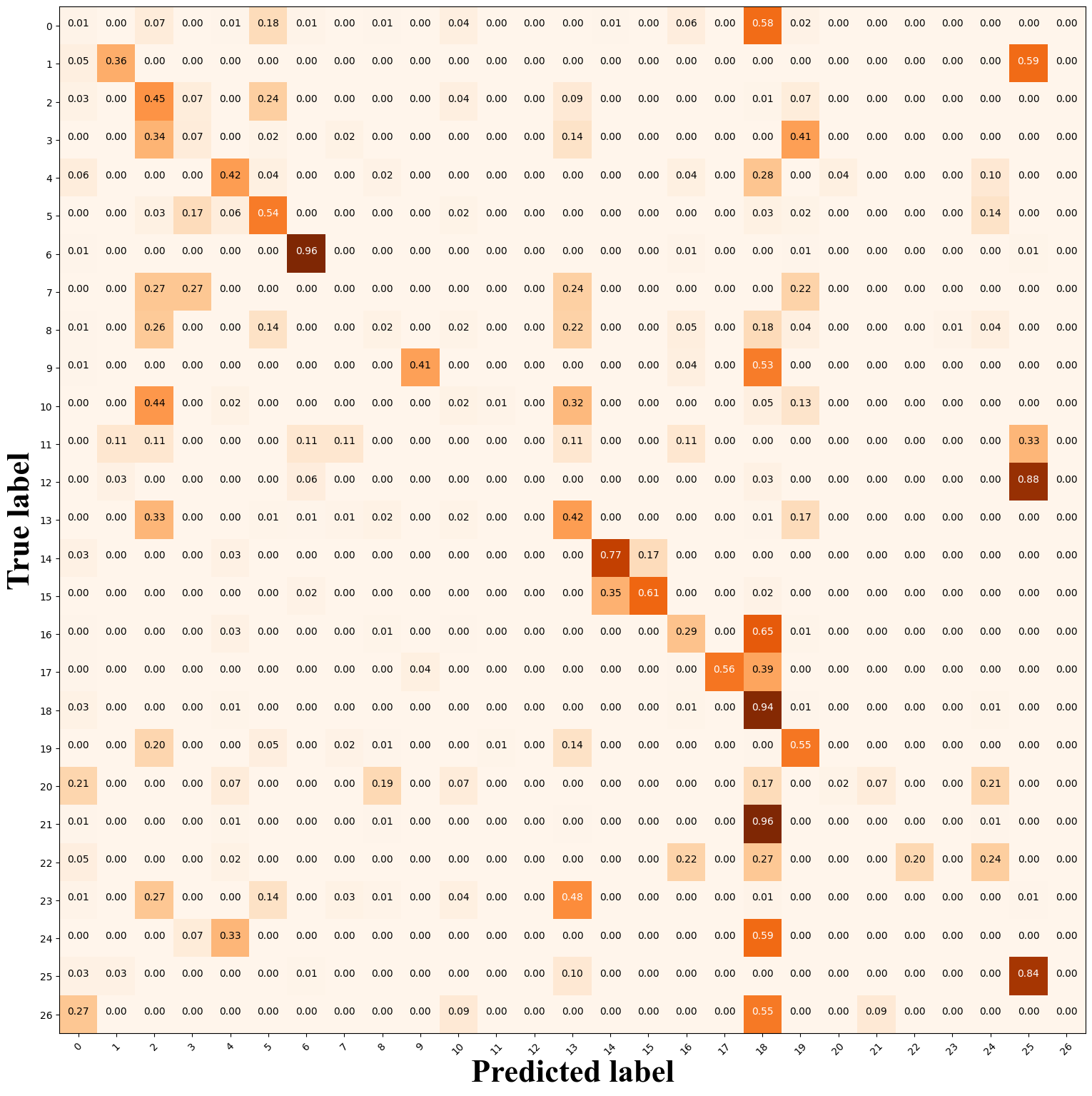}
    
    \caption{Confusion matrix for activity classification using our method with GPT-4o model in MuRAL.}
    \label{fig:confusion}
\end{figure}

\subsubsection{Resident Assignment}

To investigate the accumulation and propagation of errors in LLM-based multi-resident tracking as the evaluation sequence length increases, we present resident assignment accuracy under different evaluation window sizes in ~Figure~\ref{fig: assign_vs_wins}. As a baseline, we use the “always same resident” approach, which always predicts the same resident (“A”) for each window. We observe that, for all methods and models, accuracy declines linearly as the window size (log-scaled) increases, indicating that prediction errors accumulate and propagate over longer sequences, resulting in reduced consistency.

When the window size does not exceed 64, most LLM-based methods consistently outperform the “always same resident” baseline, except our method with 4o-mini. Comparing models, both LAHAR and our method show that the 4o model performs better than the 4o-mini model in subject assignment, highlighting the importance of strong base models for this task. Notably, LAHAR substantially improves assignment accuracy for the 4o-mini model, suggesting that input preprocessing with prior knowledge can enhance the performance of weaker models. However, for the stronger 4o model, LAHAR offers no improvement and even slightly underperforms compared to the unprocessed approach. We believe this is because LAHAR merges repeated or adjacent events from the same sensor and assumes they belong to the same resident, which does not always hold true in the complex MuRAL scenarios, leading to some misassignments.

\subsubsection{Action Description} 

For the fine-grained action-level natural language description task, our method outperforms LAHAR on both models. This is mainly because LAHAR generates descriptions by pairing sensor activation and deactivation events within a single sentence, increasing semantic granularity but deviating from the MuRAL annotation style, which describes each event separately. Consequently, LAHAR’s outputs are generally less similar to the reference annotations than those from the baseline, which aligns better with the dataset’s annotation strategy. Nevertheless, LAHAR’s approach does reduce redundancy and improves generation efficiency by merging paired events. We also observe that, for both our method and LAHAR, the 4o-mini model consistently produces descriptions that are slightly closer to the ground-truth annotations than the full 4o model. Our analysis suggests that this is due to the 4o-mini model’s stronger instruction-following capabilities, resulting in fewer omissions. Overall, regardless of the model or method, generating accurate and expressive action-level descriptions solely from ambient sensor data remains highly challenging, with significant room for improvement.

\subsubsection{Activity Classification}

For activity classification, the 4o model consistently outperforms the 4o-mini model under both the LAHAR and baseline methods. While LAHAR improves the results for the 4o-mini model, it offers little advantage compared to the larger 4o model. Overall, neither approach achieves high-accuracy zero-shot activity recognition on MuRAL.

Figure~\ref{fig:confusion} presents the confusion matrix of the best-performing 4o model, highlighting the main challenges for activity recognition on MuRAL. A major issue is that many activities—such as playing video games, watching TV, reading, and snacking—are frequently misclassified as ``resting on couch or chairs''. This misclassification arises because these activities require the model to integrate sequences of sensor events over time. For example, accurately recognizing ``playing video games'' involves tracking a resident as they turn on the TV and game console and then sit on the couch, interpreting these related actions as a single activity. Current models often fail to associate such temporally linked actions to the same individual or treat each event in isolation, leading to only the game console activation being labeled as ``playing games'', while subsequent couch-related events during gameplay are classified as ``resting on the couch''. Additionally, interactions between residents are often overlooked. For instance, when one resident is watching TV and another joins them on the couch, the model typically predicts the latter as merely resting, rather than recognizing that both residents are now watching TV together.

\section{Conclusion}

This paper presents MuRAL, a multi-resident smart-home dataset annotated with natural language, designed to explore the potential of LLMs for human activity recognition. MuRAL features fine-grained action descriptions, rich contextual information, and dynamic multi-user scenarios. The experiments on subject assignment, action description, and activity classification show that LLMs produce interpretable results but still face challenges in multi-user disambiguation and context reasoning. MuRAL provides a valuable benchmark for language-driven and socially aware HAR systems, paving the way for future research integrating multimodal inputs and interactive learning.

\section*{Ethics Statement}
The MuRAL dataset was collected under strict adherence to European General Data Protection Regulation (GDPR) and French data privacy laws (Loi Informatique et Libertés). All participants provided informed written consent for the collection and use of sensor and video data for research purposes. Video recordings were used solely for annotation purposes and were irreversibly deleted after data annotation. The final dataset is fully anonymized, contains no personally identifiable information, and was approved by the Ethics Committee of Université Grenoble Alpes (CERGA). The dataset is shared with the research community in full compliance with ethical and legal standards.

\section*{Acknowledgement}
We would like to express our sincere gratitude to Mélissa COURLA and Sybille CAFFIAU for their invaluable support and assistance in setting up and managing the DOMUS testbed. We are also deeply thankful to all the volunteers who generously participated in the data collection sessions.

%
%
\bibliographystyle{splncs04}
\bibliography{biblio}

\end{document}